# Deriving a Generalized, Actuator Position-Independent Expression for the Force Output of a Scissor Lift

Amay Saxena
University of California, Berkeley
Email: amaysaxena@berkeley.edu

***Abstract-***

*Scissor lifts, a staple of mechanical design, especially in competitive robotics, are a type of linkage that can be used to raise a load to some height, when acted upon by some force, usually exerted by an actuator. The position of this actuator, however, can affect the mechanical advantage and velocity ratio of the system. Hence, there needs to be a concrete way to analytically compare different actuator positions. However, all current research into the analysis of scissor lifts either focusses only on the screw jack configuration, or derives separate force expressions for different actuator positions. This, once again, leaves the decision between different actuator positions to trial and error, since the expression to test the potency of the position can only be derived once the position is chosen. This paper proposes a derivation for a general force expression, in terms of a few carefully chosen position variables, which can be used to generate the force expression for any actuator position. Hence, this expression illustrates exactly how each of the position variables (called a, b and i in this paper, as defined later) affect the force output, and hence can be used to pick an appropriate actuator position, by choosing values for the position variables that give the desired result.*

## I. Introduction

Scissor lifts, as depicted in figure 1, are a type of mechanism that allows for vertical displacement of some load, through the use of linked, folding supports, in a crisscross "X" pattern, referred to as a pantograph (or, simply, a scissor mechanism). Scissor lifts are widely used in industrial applications, and also form a staple design element in competitive robotics. Each arm of the crosses is called a 'scissor arm' or 'scissor member'. The upward motion is produced by the application of force, by some actuator (usually hydraulic, pneumatic, or mechanical), to the outside of the one set of supports, elongating the crossing pattern, and propelling the load vertically. However, the positioning of the actuator, in terms of the point of application of the force on the pantograph, can affect the force required of the actuator for a given load. Prudent placement of the actuator can greatly reduce the force required and the stress levels in the adjacent scissor arms.

So far, all literature on the force analysis of scissor lifts relies on an actuator-position dependent approach, where a different force expression is derived for every new actuator position. This has clear drawbacks, such that when a new actuator position is to be implemented, a new expression must be derived from first principles. This research aims to derive a generalized force equation, which may be implemented for any actuator position, with the adjustment of a few position variables. The method outlined by Spackman, 1989, of deriving a force equation as a function of the derivative of scissor height with respect to actuator length is used here as a starting point.

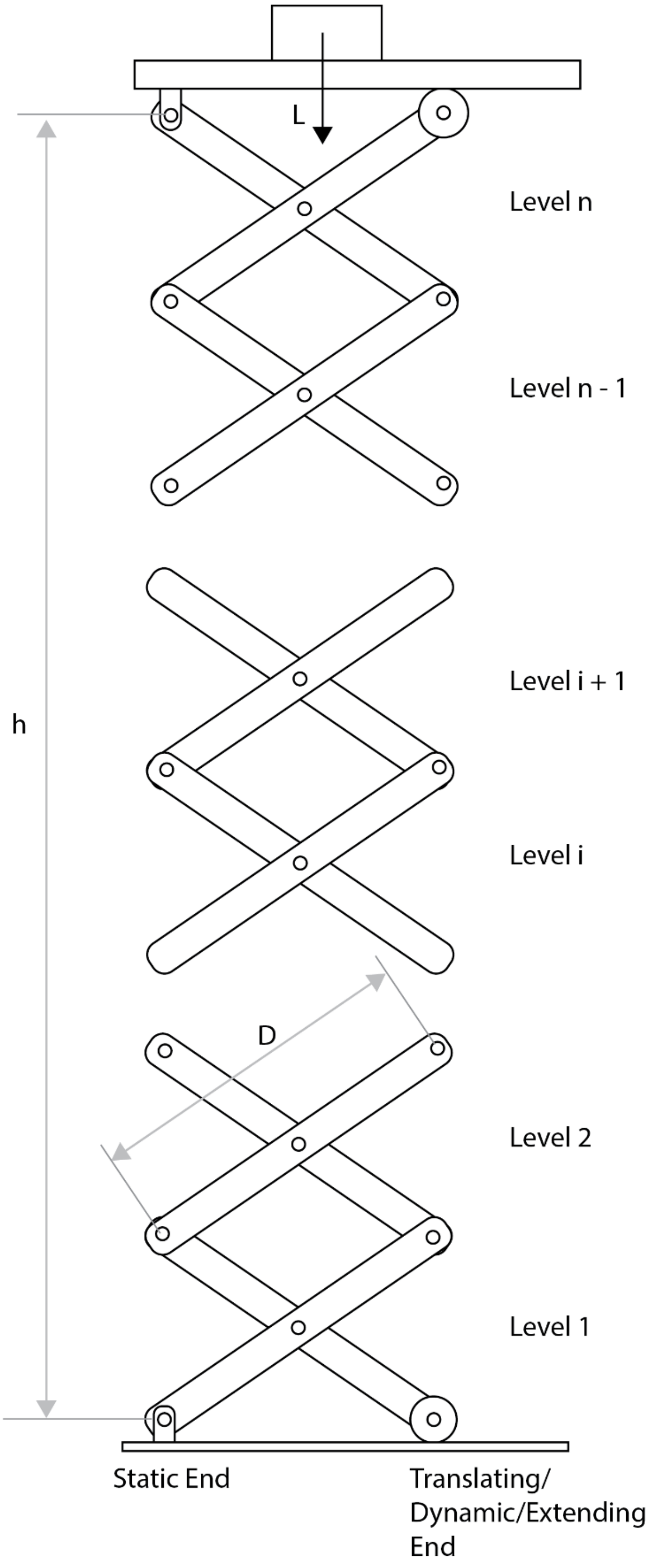

*Figure 1- An n-stage scissor lift; an illustration of the terminology used in this paper*

For the purposes of this paper, we will be deriving and using an equation for the output force of the actuator as a function of the rate of change of height with respect to actuator length. This was demonstrated as a more potent method of force calculation by Spackman in 1989 (Spackman H. , 1989). For the sake of completeness, the entire derivation of this form has been included.

The force equation fill be of the form:

$$F = f\left(\frac{dh}{dl}\right)$$

## II. Accounting for Mass of Scissor Lift

The scissor lift has a non-negligible mass. A significant amount of work will be done by the actuator in lifting the mass of the scissor lift itself to any given height $h$. Hence, this work must be accounted for. This section 2.1 will derive an equation for the work done in lifting the weight of the scissor lift to any height $h$.

We will model the mass of the scissor lift in terms of an arbitrary cuboidal mass $m$, of weight $B = mg$, and dimensions $a \times b \times h$. (Fig 1.0.) (B is used to avoid confusion with any other variables)

We will examine the behaviour of the mass as the height increases, but mass remains the same (as is the case with a scissor lift in extension). For the sake of simplicity, the density ($\rho$) is taken to be constant (although this is not the case)

Consider an infinitesimal slice of this solid, of thickness $dy$, at a height $y$ from the base of the solid.

The mass of this slice then becomes:

$$ab(dy)\rho$$

At a height $y$, the potential energy of the slice is:

$$ab\rho y(dy)$$

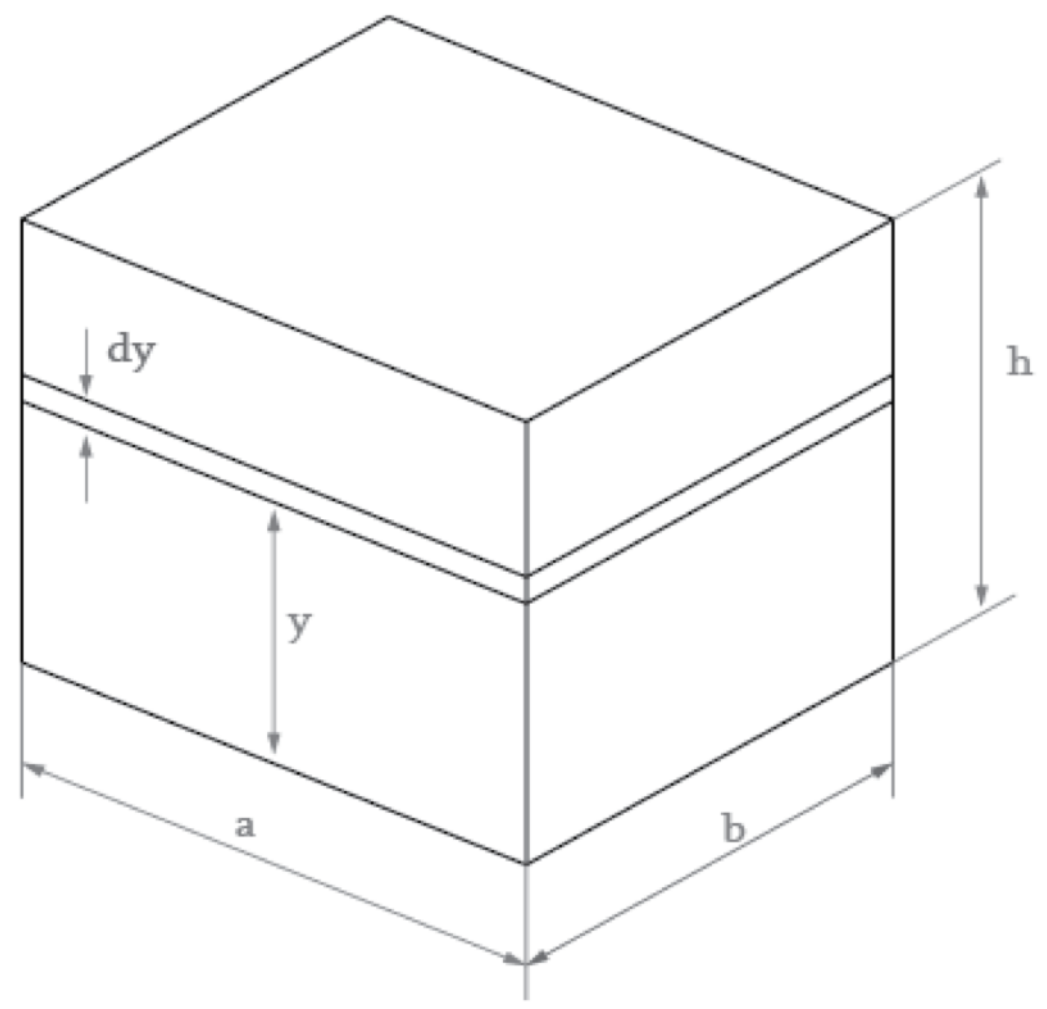

Figure 2- Arbitrary mass

Thus, the potential energy ($W_B$) of the entire mass of height $h$ can be approximated by

$$W_B = \int_{y=0}^{y=h} ab\rho y\,(dy)$$

We know that weight of mass, B is:

$$B = abh\rho$$

$$\Rightarrow ab\rho = \frac{B}{h}$$

$$\Rightarrow W_B = \int_{y=0}^{y=h} \frac{B}{h} y(dy)$$

$$\Rightarrow W_B = \frac{B}{h}\left[\frac{y^2}{2}\right]_{y=0}^{h}$$

$$\Rightarrow W_B = \frac{B}{h}\frac{h^2}{2} = \frac{Bh}{2}$$

$$\Rightarrow W_B = \frac{B}{2}h \qquad (i)$$

Now, if the height of the block changes but weight remains the same, work done in changing height, $\Delta W_B$ is given by:

$$\Delta W_B = W_2 - W_1$$

$$\Rightarrow \Delta W_B = \frac{B}{2}(h_2 - h_1)$$

Work done is the product of force and displacement. In the above equation (i), displacement is $h_2 - h_2$, and force is $\frac{B}{2}$, which shows that the work done in keeping the weight of this arbitrary mass at height $h_2$, is the work done in bringing half its weight to height $h_2$.

Hence, the weight of the scissor lift can be modelled by placing half its weight at the top, i.e. by adding half of the weight of the scissor lift to the load.

Hence, for the rest of this paper, the weight of the lift will be accounted for by taking the effective load $L_E$ , for a particular load L, such that:

$$\boldsymbol{L_E = L + \frac{B}{2}} \qquad (\boldsymbol{ii})$$

## III. Deriving a General Equation for Force Required

Let the force output of the actuator, as a function of time, be $F$, and length of actuator be $l$

Applying conservation of energy, we have

$$Work\ done\ by\ actuator = Work\ done\ to\ take\ lift\ to\ height\ h$$

$$\Rightarrow \int_{l_1}^{l_2} F\, dl = \int_{h_1}^{h_2} L_E\, dh$$

$$\Rightarrow \int_{l_1}^{l_2} F\, dl = \int_{h_1}^{h_2} \left(L + \frac{B}{2}\right) dh$$

Differentiating both sides with respect to actuator length $l$ gives the required force equation:

$$\boldsymbol{F = \left(L + \frac{B}{2}\right) \frac{dh}{dl}} \qquad (1)$$

(1) will be used to calculate the theoretical force required to raise a particular load L, for a particular actuator placement.

The rest of this paper will focus on deriving a generalized expression for $\frac{dh}{dl}$, that may be implemented for any actuator position. This will then be substituted into (1) to obtain the required generalized force equation.

Note that this implies that mechanical advantage $= \frac{F}{L_E} = \frac{dh}{dl}$

# IV. The Actuator

This paper will discuss the force applied on the scissor lift in terms of a linear actuator, in which one end translates (extends out) and does work on the lift.

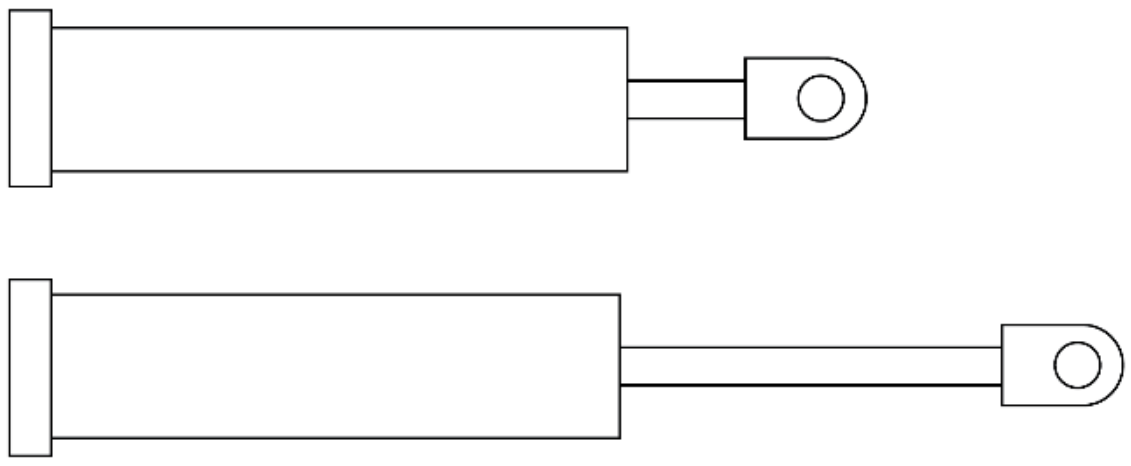

*Figure 3- Linear actuator before and after extending. When the actuator extends, it does work.*

Usually, the translating end will be attached to some point on the lift, and the static end will be attached to some fixed support. Hence, when the actuator extends, it will cause the lift to extend.

It should be noted that this does not imply loss of generality, as the derivation holds true even if the force is applied by any means other than a linear actuator.

Fig. 3 is an example of the use of a linear actuator in a 2-stage scissor lift.

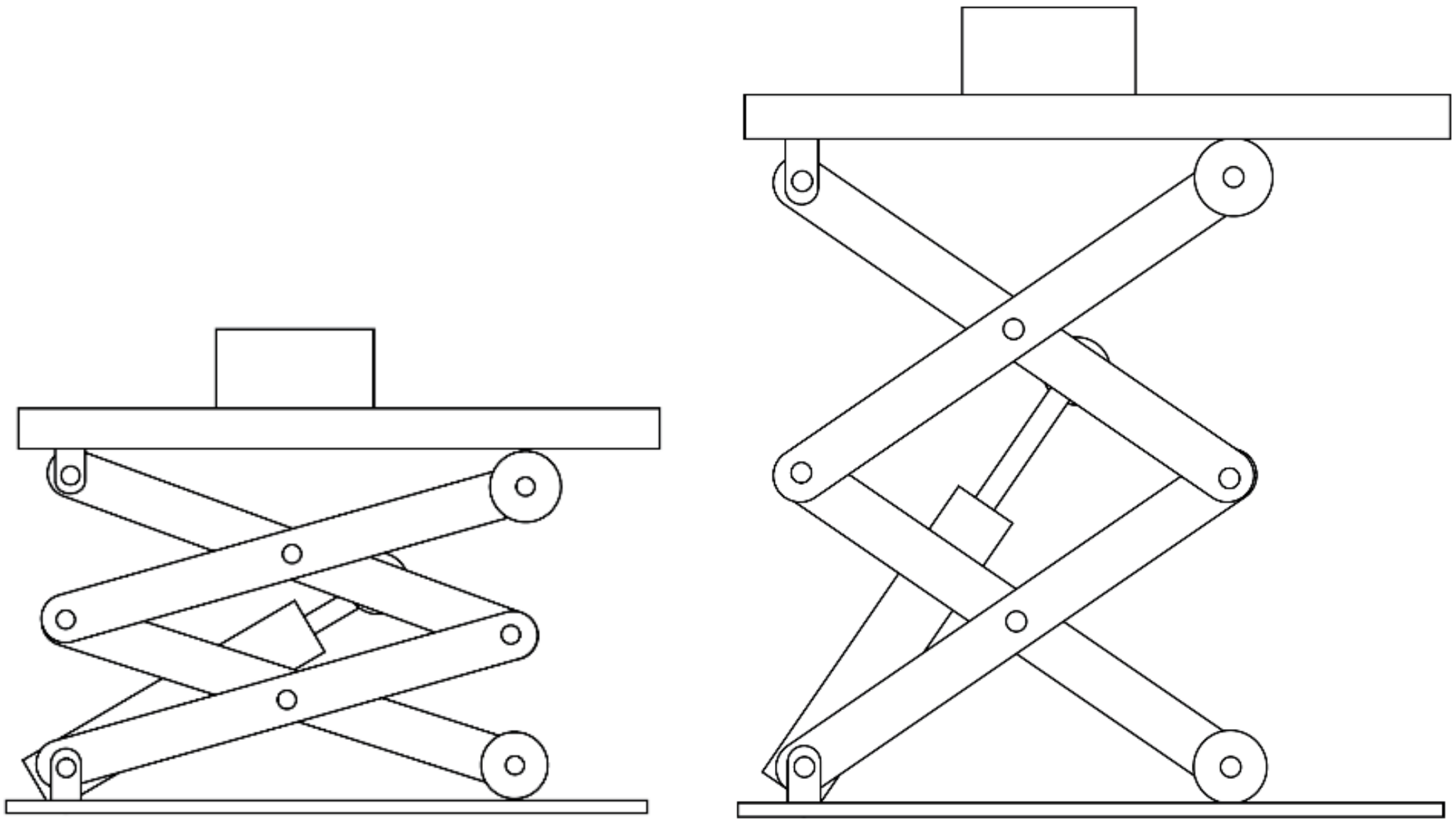

*Figure 4- Example of a 2-stage scissor lift powered by a linear actuator, before and after extending. In the diagram, the right side of the lift is free to move, and is hence the translating or 'extending' side and the left side is fixed, and is hence the static side.*

## V. The Derivation

First, note that the force expression must be different for when the force is applied to the mobile end of the lift, than when it is applied to the stationary end of the lift (Fig. 2)

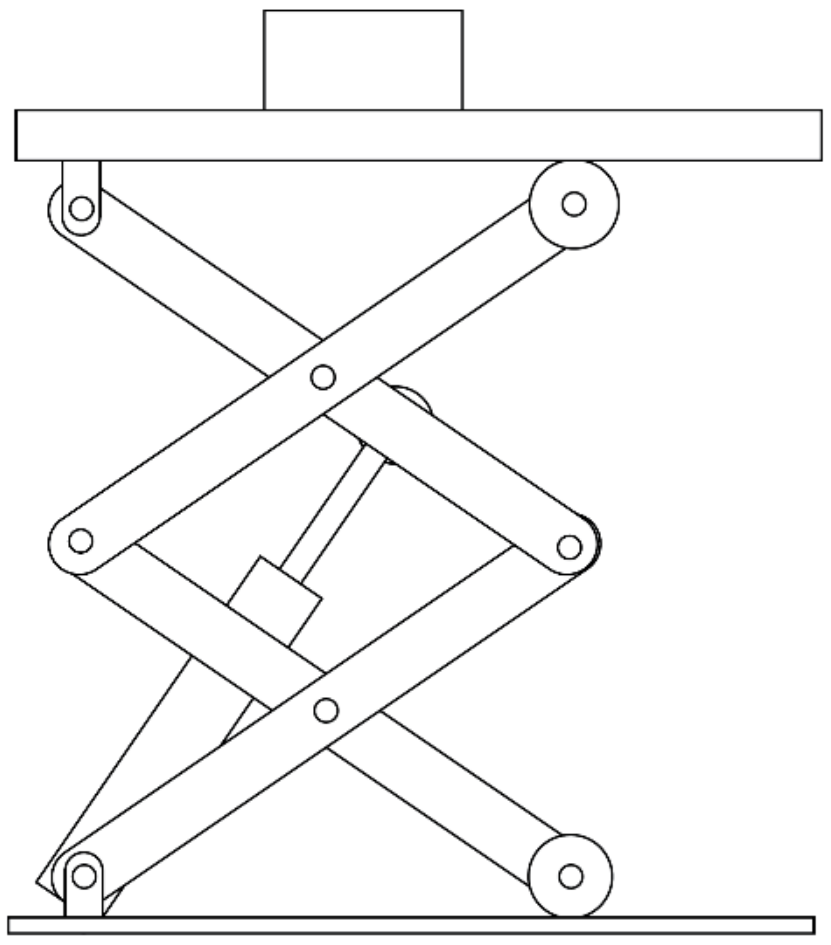

*Figure 4(a)- Force applied on mobile end*

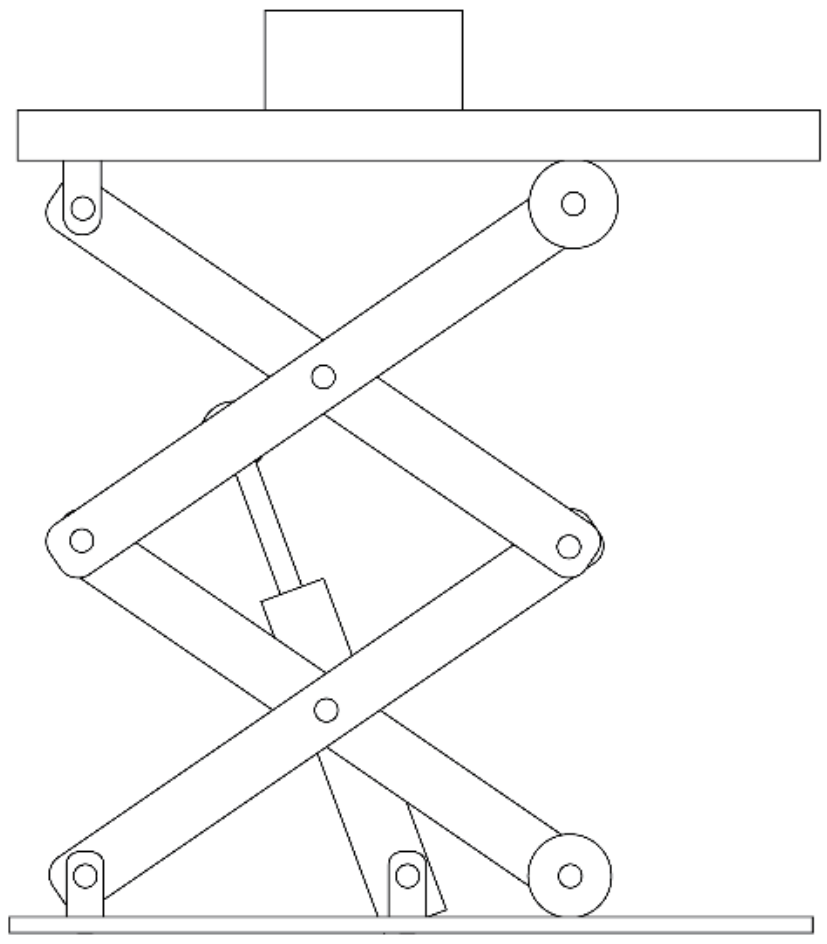

*Figure 5(b)- Force applied on static end*

These two cases must be considered separately. Case 1 shall be where force is applied by the actuator on any arm on the stationary side of the scissor lift, and Case 2 shall be where force is applied on any arm on the static side of the scissor lift.

### *A. Derivation for Case 1*

We will first consider the lift if it were embedded in a Cartesian plane, with origin at the lower most hinge on the static end of the lowermost level of the scissor lift (Fig. 5, Fig. 6). It is apparent that this point is the only point on the lift that does not translate when the lift extends, making it an appropriate choice for the origin.

We may now amend our definitions for Cases 1 & 2 slightly. We shall now define Case 1 as those actuator positions in which the actuator applies force on a negatively sloping arm of the lift, in the context of the Cartesian plane. i.e. a scissor arm that would have a negative gradient, if it were a line in this Cartesian plane, as defined above. Case 2, then, would be those actuator positions in which the actuator applies force on a positively sloping scissor arm. (See Fig. 5)

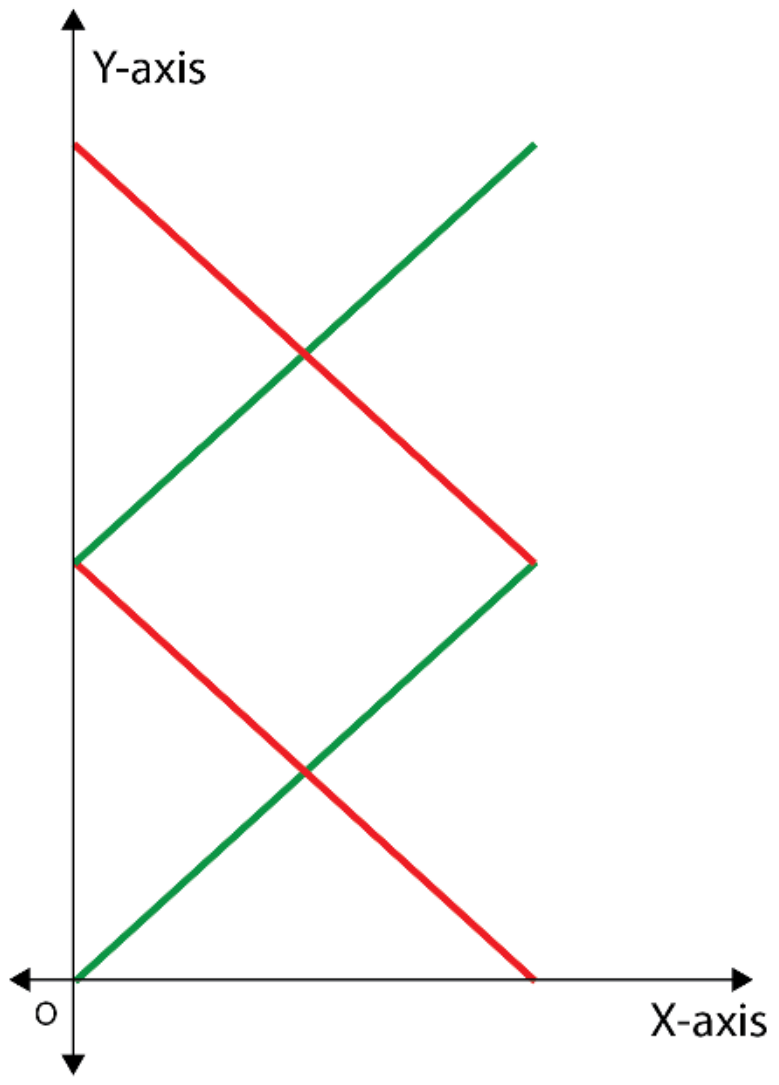

*Figure 6- Green- positively sloping arms; Red- negatively sloping arms*

Without loss of generality, consider the following arbitrary case for when force is applied on a positively sloping arm of the scissor lift:

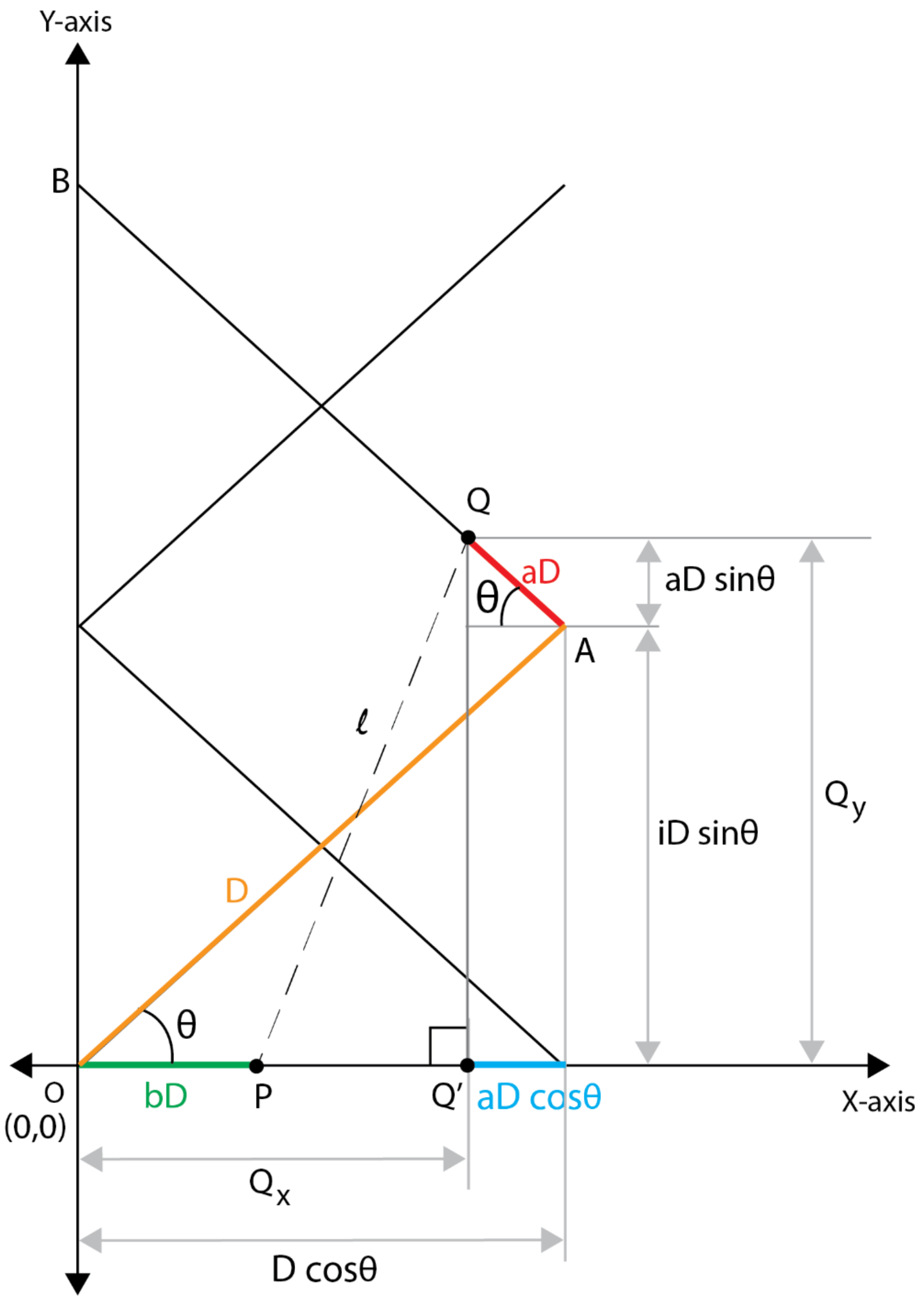

*Figure 7- General instance of Case 1; Note that although the diagram shown represents a 2-stage scissor lift, the derivation that follows addresses the fact that the list could have any number of stages.*

In Fig. 6 above;

Let Q be the point at which force is applied (i.e. the extending end of the actuator is attached). Let P be the point at which the static end of the actuator is attached. Line PQ is hence the actuator. Length of line PQ is the actuator length.

Let $Q'$ be the projection of point $Q$ on the X-axis.

O is the origin, as described earlier.

$\theta$ is the angle between any of the positively sloping scissor arms and the X-axis.

$h$ is the height that the load has been lifted to.

$i$ is the number of complete scissor levels immediately below point Q.

(In Fig. 4 $i = 1$, i.e. there is only 1 complete scissor level below point Q, but the derivation holds true for any value of i)

$l$ is the length of the actuator, $\overline{PQ}$

D is the length of the scissor arms.

$\overline{AQ} = aD \quad |0 < a < 1$

i.e. Let the distance between point Q and the hinge on the mobile side of the scissor lift, on the same arm as Q, be $aD$, where $0 < a < 1$ . i.e. $a$ is the fraction that AQ is of the full length of the arm AB.

Similarly,

Let $\overline{OP} = bD \quad |b \in \mathbb{R}$

Let $n$ be the number of levels of the scissor lift.

First, note that

$$h = nD\ sin\theta \tag{2}$$

Since the scissor arm of length D at angle $\theta$ will have a vertical height of $D\ sin\theta$, and the total height $h$ will be by virtue of $n$ such levels.

Now consider the co-ordinates of point Q, relative to origin O, as $(Q_x\ , Q_y)$

From Fig. 6, it is clear that:

$$Q_x = (1 - a)D\ cos\theta \tag{3}$$

$$Q_y = (i + a)D\ sin\theta \tag{4}$$

In Fig. 6, consider the right angled triangle where the actuator is the hypotenuse, $\Delta PQQ'$

In $\Delta PQQ'$,

$$\overline{PQ} = l \tag{5}$$

$$\overline{QQ'} = Q_y \tag{6}$$

$$\overline{PQ'} = Q_x - bD \tag{7}$$

Substituting (3) into (7) and (4) into (6), we have:

$$\overline{QQ'} = (i + a)D\ sin\theta \tag{8}$$

$$\overline{PQ'} = (1 - a)D\ cos\theta - bD \tag{9}$$

Applying the Pythagorean Theorem to $\Delta PQQ'$:

$$\overline{PQ}^2 = \overline{QQ'}^2 + \overline{PQ'}^2 \tag{10}$$

Substituting (5), (8) and (9) into (10):

$$l^2 = (i+a)^2 D^2 \, sin^2\theta \; + \; D^2[(1-a)\, cos\theta - b]^2 \tag{11}$$

$$\Rightarrow \quad l^2 = (i+a)^2 D^2 \, sin^2\theta \; + \; D^2[(1-a)^2 \, cos^2\theta - 2b(1-a)\cos\theta + b^2] \tag{12}$$

$$\Rightarrow \quad \frac{l^2}{D^2} = [(1-a)^2 - (i+a)^2]\, cos^2\theta \; - 2b(1-a)\cos\theta + b^2 + (i+a)^2 \tag{13}$$

Since $l/D$ is always positive, we may take a square root on both sides:

$$\frac{l}{D} = \sqrt{[(1-a)^2 - (i+a)^2]\, cos^2\theta \; - 2b(1-a)\cos\theta + b^2 + (i+a)^2} \tag{14}$$

Differentiating with respect to $h$:

$$\frac{1}{D}\frac{dl}{dh} = \frac{1}{2}\frac{2b(1-a)\, sin\theta \,\frac{d\theta}{dh} - [(1-a)^2 - (i+a)^2]\, 2\, cos\theta \, sin\theta \,\frac{d\theta}{dh}}{\sqrt{[(1-a)^2 - (i+a)^2]\, cos^2\theta \; - 2b(1-a)\cos\theta + b^2 + (i+a)^2}} \tag{15}$$

To proceed, we must find an expression for $\frac{d\theta}{dh}$.

From (2), we have:

$$h = nD \, sin\theta \tag{2}$$

Differentiating (2) with respect to $\theta$ gives:

$$\frac{dh}{d\theta} = nD \, cos\theta \tag{16}$$

Reciprocate (18) to give the required expression for $\frac{d\theta}{dh}$

$$\frac{d\theta}{dh} = \frac{1}{nD \, cos\theta} \tag{17}$$

Substituting (19) in (17):

$$\frac{dl}{dh} = \frac{1}{n} \cdot \frac{b(1-a)\tan\theta - [(1-a)^2 - (i+a)^2]\sin\theta}{\sqrt{[(1-a)^2 - (i+a)^2]\, cos^2\theta \; - 2b(1-a)\cos\theta + b^2 + (i+a)^2}} \tag{18}$$

Reciprocate to give the required expression for $\frac{dh}{dl}$

$$\frac{dh}{dl} = n \cdot \frac{\sqrt{[(1-a)^2 - (i+a)^2]\, cos^2\theta \; - 2b(1-a)\cos\theta + b^2 + (i+a)^2}}{b(1-a)\tan\theta - [(1-a)^2 - (i+a)^2]\sin\theta} \tag{19}$$

Substitute (23) in (1) to obtain the final force equation for Case 1:

$$F = n\left(L + \frac{B}{2}\right)\frac{\sqrt{[(1-a)^2 - (i+a)^2]\, cos^2\theta\; - 2b(1-a)\cos\theta + b^2 + (i+a)^2}}{b(1-a)\tan\theta - [(1-a)^2 - (i+a)^2]\sin\theta} \tag{20}$$

In order to simplify the expression, we will define the following constants for any given actuator position:

$$A = 1 - a \tag{21}$$

$$B = i + a \tag{22}$$

To generate the force expression for any actuator position, first obtain the constants A and B for that actuator position through (25) and (26), then substitute in the final equation:

$$F = n\left(L + \frac{B}{2}\right)\frac{\sqrt{(A^2 - B^2)\, cos^2\theta - 2bA\cos\theta + b^2 + B^2}}{bA\tan\theta - (A^2 - B^2)\sin\theta} \tag{23}$$

An alternate form involves defining three variables, A, B, and C, such that

$$A = (1-a)^2 - (i+a)^2 \tag{24}$$

$$B = b(1-a) \tag{25}$$

$$C = b^2 + (i+a)^2 \tag{26}$$

Here, the final expression becomes:

$$F = n\left(L + \frac{B}{2}\right)\frac{\sqrt{A\, cos^2\theta - 2B\cos\theta + C}}{B\tan\theta - A\sin\theta} \tag{27}$$

### *B. Derivation for Case 2*

Using the same approach as in Case 1, we generate the following figure, with each point having the same definition as in Fig. 4

Using the same definition for Q and its co-ordinates, from Fig. 7, we have:

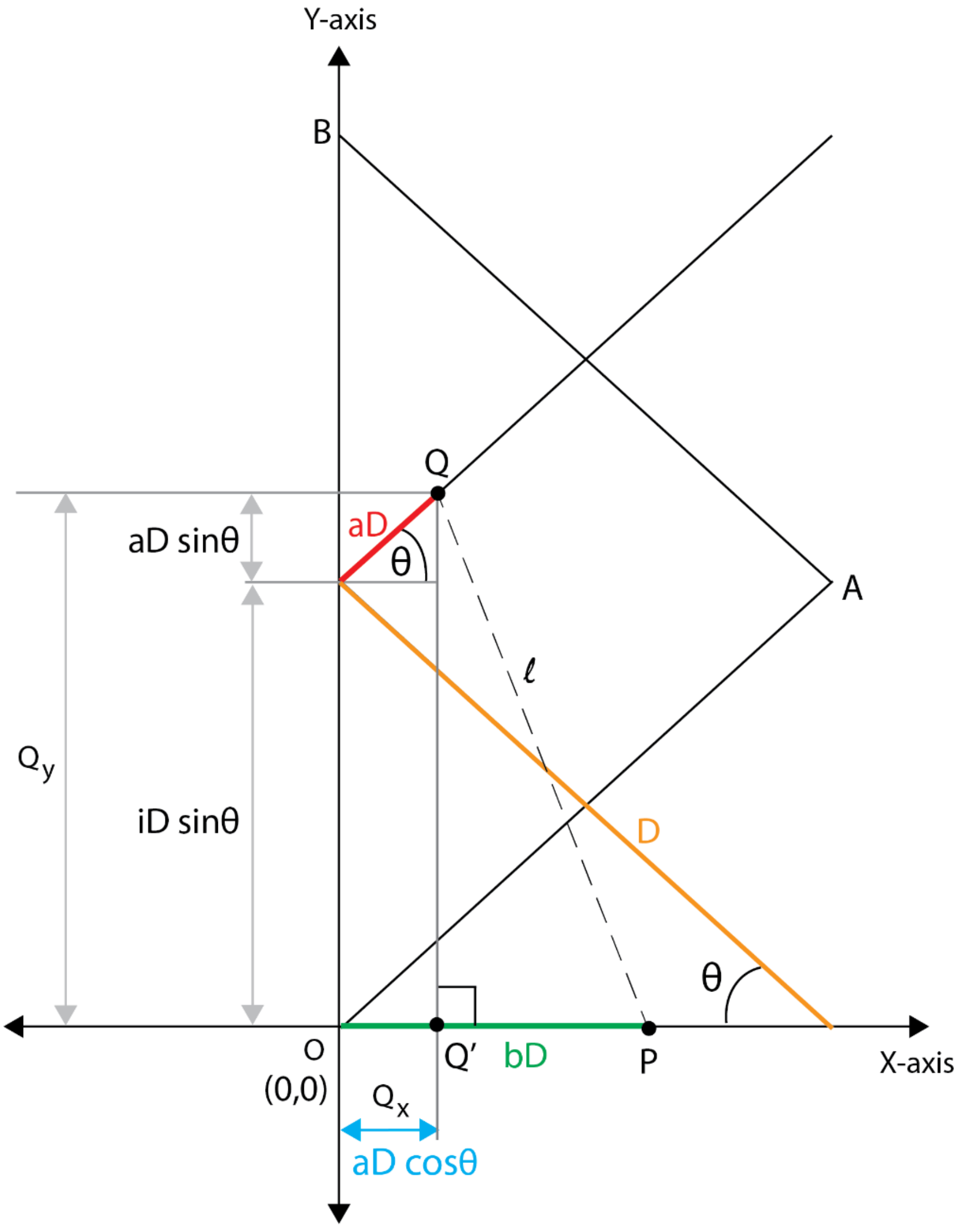

*Figure 8- General instance of Case 2. Definitions of all points remains the same as in Fig. 6*

$$Q_x = aD\ cos\theta \quad (28)$$
$$Q_y = (i + a)D\ sin\theta \quad (29)$$

As in Case 1, point $Q'$ is the projection of point $Q$ on the X-axis.

In Fig. 7, consider right angled triangle, $\Delta QQ'P$

In $\Delta QQ'P$,

$$\overline{PQ} = l \quad (30)$$
$$\overline{QQ'} = Q_y \quad (31)$$

$$\overline{PQ'} = bD - Q_x \tag{32}$$

Substituting (25) into (29) and (26) into (28):

$$\overline{QQ'} = (a + i)D\ sin\theta \tag{33}$$

$$\overline{PQ'} = D(b - a\ cos\theta) \tag{34}$$

Using the same approach as in Case 1, Applying the Pythagorean Theorem on $\Delta PQQ'$, and then using the substitutions in (34), (37) and (38):

$$l^2 = D^2(a + i)^2\ sin^2\theta + D^2(b - a\cos\theta)^2 \tag{35}$$

$$\Rightarrow \quad \frac{dl}{dh} = \frac{1}{n} \cdot \frac{ab\tan\theta - i(2a + i)\ \sin\theta}{\sqrt{b^2 + (a + i)^2 - \ 2ab\cos\theta + i(2a + i)\ cos^2\theta}} \tag{36}$$

Reciprocate (36) to give the required expression for $\frac{dh}{dl}$

$$\frac{dh}{dl} = n\frac{\sqrt{b^2 + (a + i)^2 - \ 2ab\cos\theta + i(2a + i)\ cos^2\theta}}{ab\tan\theta - i(2a + i)\ \sin\theta} \tag{37}$$

Substitute (46) in (1) to give the force expression for Case 2:

$$F = n\left(L + \frac{B}{2}\right)\frac{\sqrt{b^2 + (a + i)^2 - \ 2ab\cos\theta + i(2a + i)\ cos^2\theta}}{ab\tan\theta - i(2a + i)\ \sin\theta} \tag{38}$$

To simplify the expression, as we did for Case 1, we retain the definitions for A, B, and C.

We must also define a new constant, D, such that:

$$D = i(2a + i) \tag{39}$$

Use the definitions of A, B, C and D, in (28), (29), (30) and (47), to obtain the simplified force equation:

$$F = n\left(L + \frac{B}{2}\right)\frac{\sqrt{C - \ 2(b - B)\cos\theta + D\ cos^2\theta}}{(b - B)\tan\theta - D\sin\theta} \tag{40}$$

# VI. Conclusion:

In order to generate the force expression for a scissor lift with any actuator position, first obtain the following constants:

$$A = (1-a)^2 - (i+a)^2$$
$$B = b(1-a)$$
$$C = b^2 + (i+a)^2$$
$$D = i(2a+i)$$

Now determine whether force is being applied on a negatively sloping arm of the lift or a positively sloping arm.

If force is applied on a negatively sloping arm:

$$F = n\left(L + \frac{B}{2}\right)\frac{\sqrt{A\,cos^2\theta - 2B\cos\theta + C}}{B\tan\theta - A\sin\theta}$$

If force is applied on a positively sloping arm:

$$F = n\left(L + \frac{B}{2}\right)\frac{\sqrt{C - \ 2(b-B)\cos\theta + D\,cos^2\theta}}{(b-B)\tan\theta - D\sin\theta}$$

### *A. A Note on Picking a, b and i*

The most important part of using this system of expressions is picking the appropriate values for $a$, $b$ and $i$, for the actuator position being tested. This depends entirely on which arm the point of application of force is considered. This is usually self-apparent, however it can get ambiguous when force is applied to a hinge that is shared by the extreme points of two arms, one positively sloping and one negatively sloping. In such a case, we will get drastically different values for all three constants, depending on which arm is selected as the one on which the point of application of force lies.

As a direct consequence of the methodology used in the derivation, in particular, the definition of the constant $a$, the arm that should be considered is the one that lies entirely above a line passing through the point of force application and parallel to the X-axis. Another way of framing this is that the point of application is *not* to be considered as being applied on an arm that is included in any of the scissor levels being counted to obtain $i$.

An application of this rule shall be seen in Sec. VII. B.

## VII. Examples and Verification:

We will now use this system to arrive at the force expressions for 2 actuator positions, for which an expression has been derived from first principles in the past. This will illustrate the fact that this system greatly reduces the effort needed to arrive at a force expression for any arbitrary actuator position.

### *A. Example Position 1*

The first position that will be considered is the most prevalent 'Screw jack' configuration, in which the force is applied as shown:

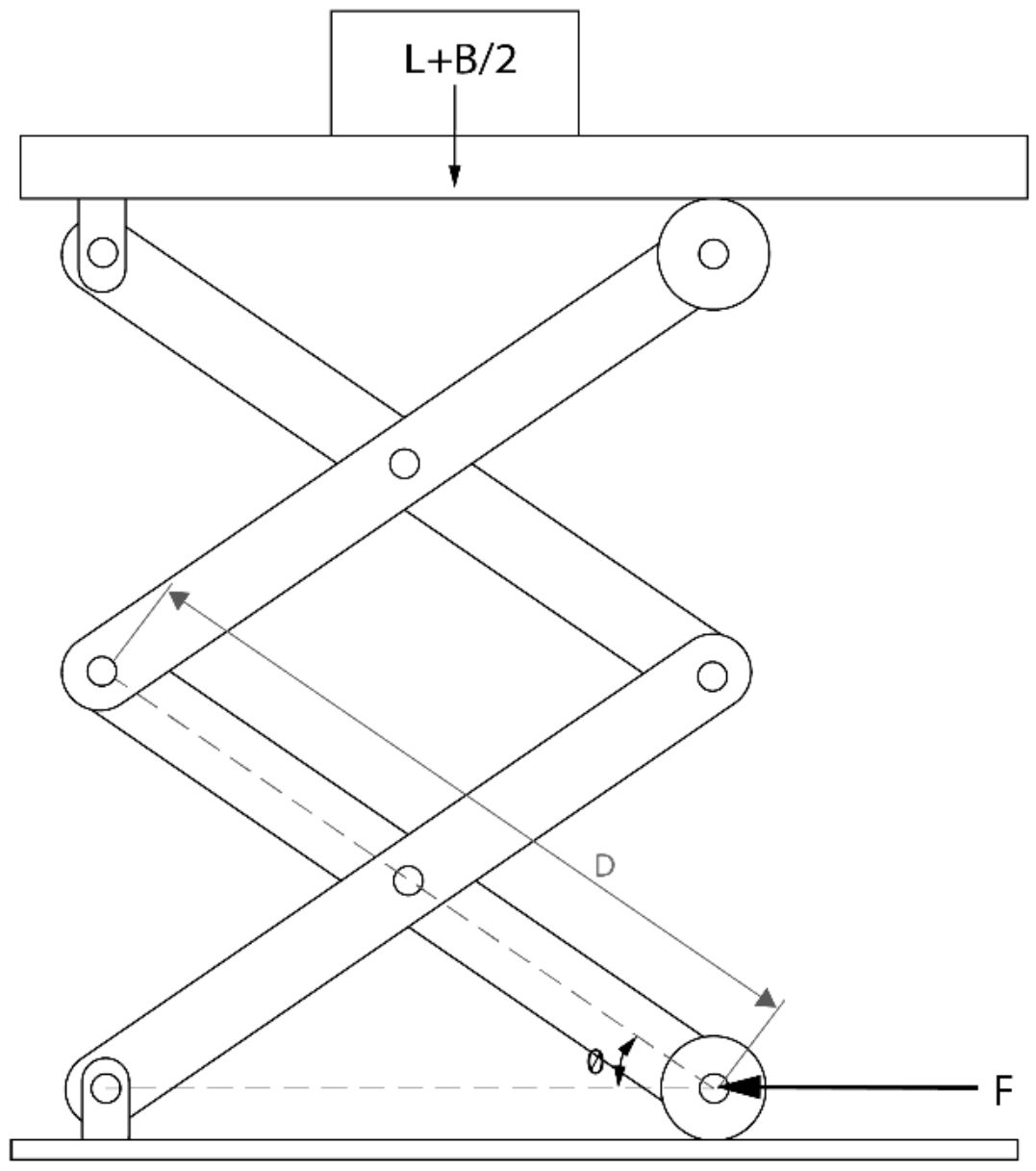

*Figure 9(a)- Free Body diagram of the scissor lift under the actuator force F.*

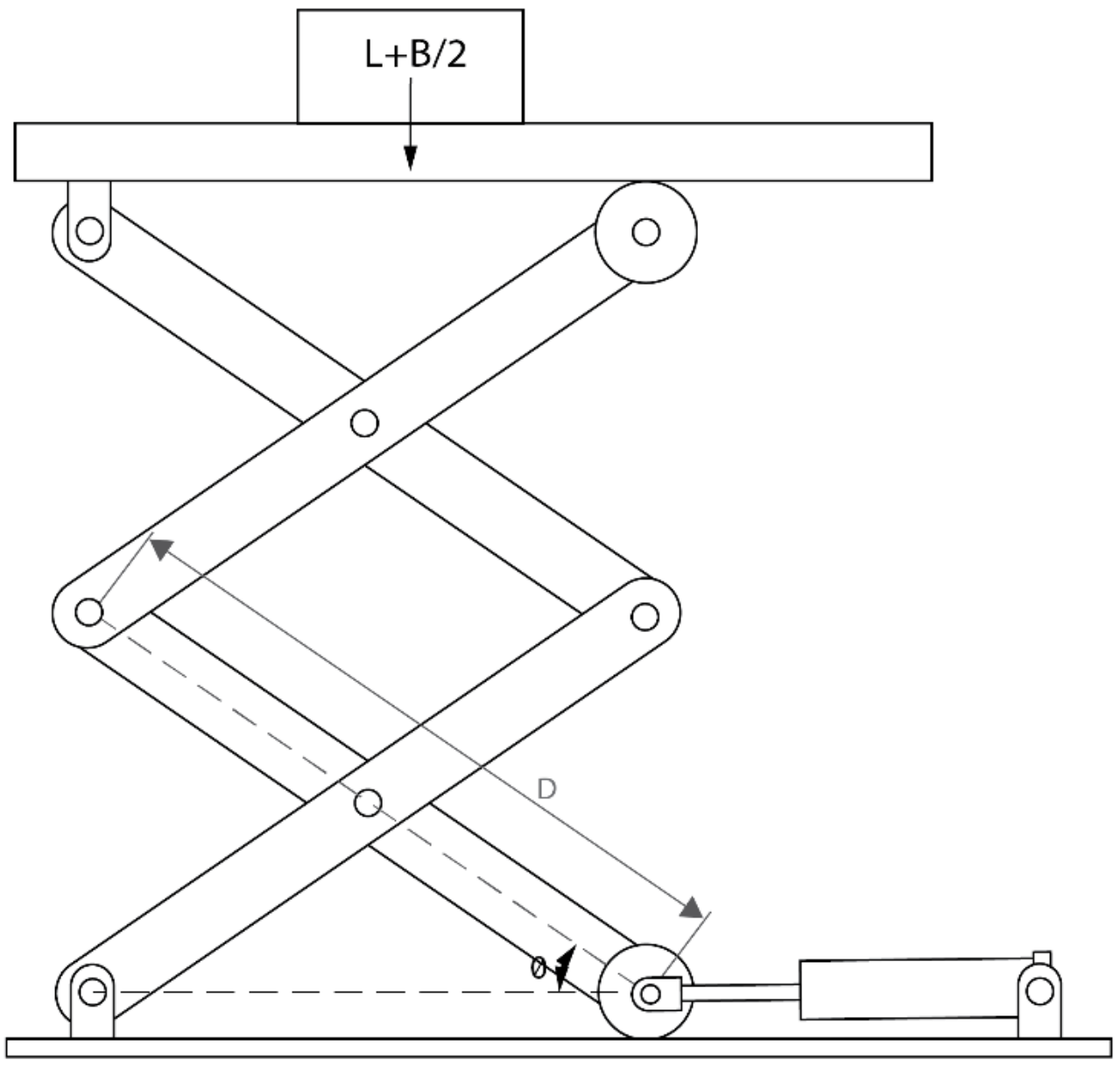

*Figure 8(b)- Using the linear actuator analogy, this is what the configuration for example 1 would look like. However, force may be applied in this direction by any means*

The known force expression for this actuator position, derived through analysis of static equilibrium and other first principles methods, is:

$$F = \left(L + \frac{B}{2}\right)\frac{n}{\tan\theta} \tag{41}$$

We will show that this result can be obtained from the system derived in his paper, in a simpler manner.

From Fig. 8(b), we see that the actuator is applying force on the first negatively sloping arm of the lift. Hence, we must use the expression for negatively sloping arms:

$$F = n\left(L + \frac{B}{2}\right)\frac{\sqrt{A\,cos^2\theta - 2B\cos\theta + C}}{B\tan\theta - A\sin\theta} \tag{42}$$

We must first obtain a, b and i, as defined earlier:

From Fig 8(b), we can see that force is being applied directly at the hinged part of the arm. Hence, the distance between the point of force application and the hinge on the same arm directly below the point, is zero.

i.e. $a = 0$

The number of complete scissor levels (complete crosses) is zero.

i.e. $i = 0$

From Fig. 8(b) we can see that depending on the length of the actuator, the distance between the origin and the static end of the actuator could be any arbitrary length. Hence, we must let b be variable, as it could take any value. We must hence show that the derivation for the force expression becomes independent of b.

Let $b = b$

We must now obtain the three constants, A, B and C, given that:

$a = 0$
$b = b$
$i = 0$

The constants are given by:

$A = (1-a)^2 - (i+a)^2 = 1 - 0 = 1$
$B = b(1-a) = b(1-0) = b$
$C = b^2 + (i+a)^2 = b^2 + 0 = b^2$

Hence, the force expression by this system becomes:

$$F = n\left(L + \frac{B}{2}\right)\frac{\sqrt{cos^2\theta - 2b\cos\theta + b^2}}{b\tan\theta - \sin\theta} \tag{43}$$

$$\Longrightarrow \quad F = n\left(L + \frac{B}{2}\right)\frac{\sqrt{cos^2\theta - 2b\cos\theta + b^2}}{b\tan\theta - \sin\theta} \tag{44}$$

$$\Longrightarrow \quad F = n\left(L + \frac{B}{2}\right)\frac{\sqrt{(\cos\theta - b)^2}}{b\tan\theta - \sin\theta} \tag{45}$$

$$\Longrightarrow \quad F = \pm\, n\left(L + \frac{B}{2}\right)\frac{\cos\theta - b}{b\tan\theta - \sin\theta} \tag{46}$$

Substituting $\tan\theta \equiv \frac{\sin\theta}{\cos\theta}$

$$\Longrightarrow \quad F = \pm\, n\left(L + \frac{B}{2}\right)\frac{\cos\theta - b}{b\frac{\sin\theta}{\cos\theta} - \sin\theta} \tag{47}$$

$$\Longrightarrow \quad F = \pm\, n\left(L + \frac{B}{2}\right)\frac{\cos\theta\,(\cos\theta - b)}{\sin\theta\,(b - \cos\theta)} \tag{48}$$

$$\Longrightarrow \quad F = \pm\, n\left(L + \frac{B}{2}\right)\frac{\cos\theta\,(\cos\theta - b)}{\sin\theta\,(\cos\theta - b)} \tag{49}$$

$$\Longrightarrow \quad F = \pm\, n\left(L + \frac{B}{2}\right)\frac{\cos\theta}{\sin\theta} \tag{50}$$

$$\Longrightarrow \quad F = \pm\left(L + \frac{B}{2}\right)\frac{n}{\tan\theta} \tag{51}$$

Since force applied in that direction cannot be negative, reject the negative expression

$$\Longrightarrow \quad F = \left(L + \frac{B}{2}\right)\frac{n}{\tan\theta} \tag{52}$$

This is exactly the expression obtained in the past by analysis of quazi-static equilibrium, and was obtained through this system in a much simpler manner.

### *B. Example Position 2*

The next position that will be considered is a rather unconventional one, where force is applied in a vertical direction:

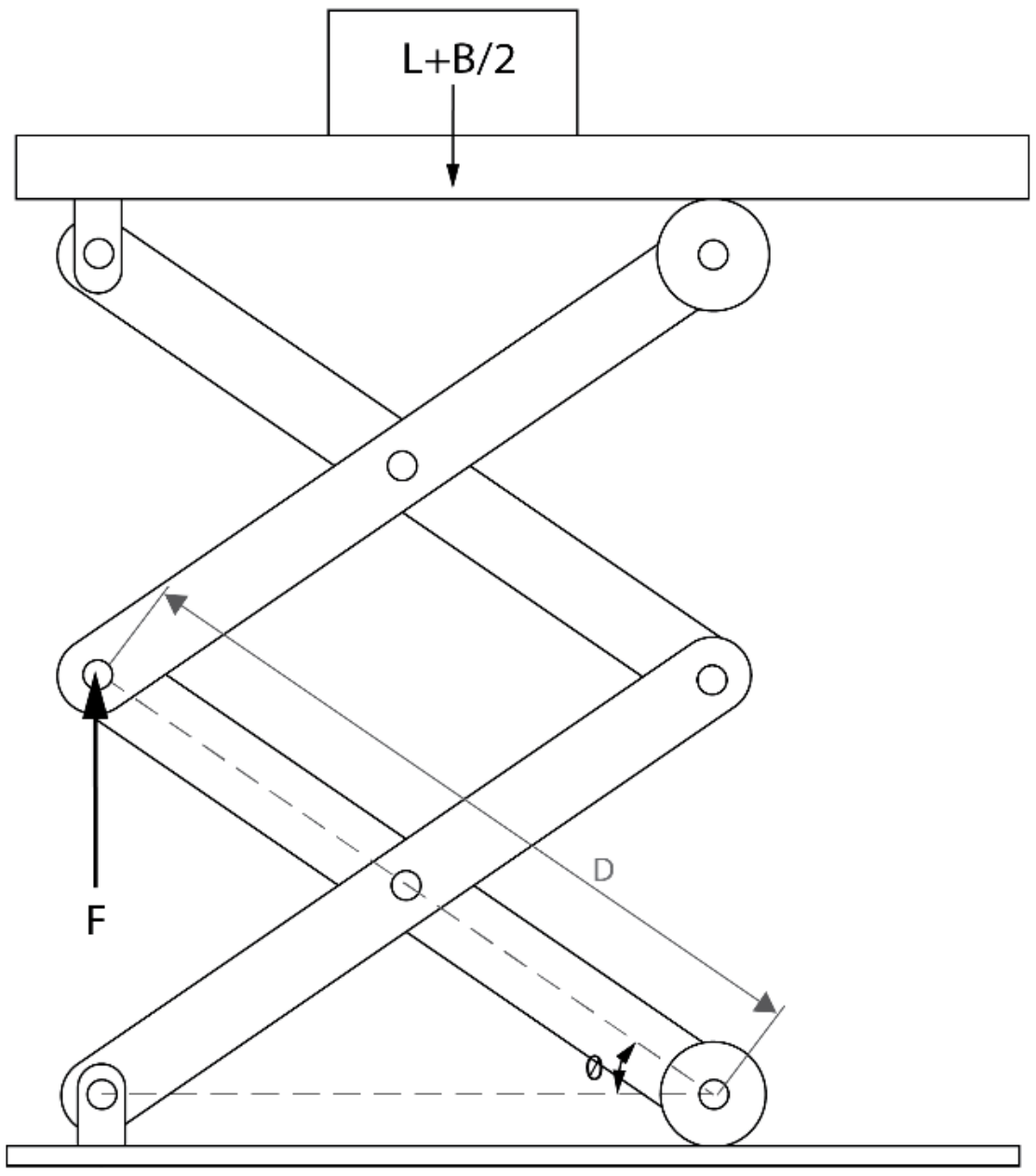

*Figure 10(a)- Free Body diagram of the scissor lift under the actuator force F.*

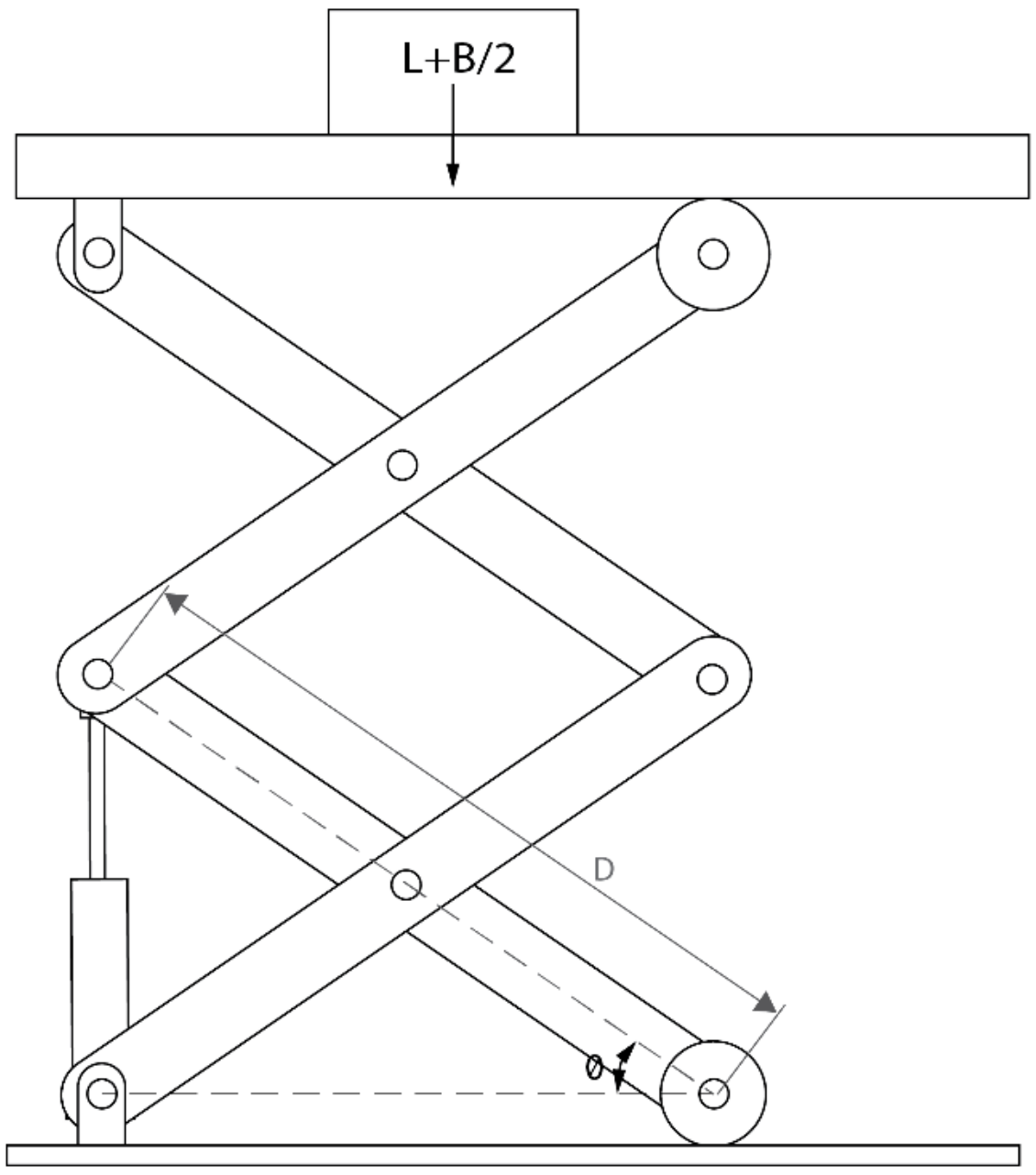

*Figure 9(b)- Using the linear actuator analogy, this is the configuration that would arise. Note however, that this is not how such an actuator position would be implemented, since at low angles, there is no space for the actuator to fit.*

From Fig, 9(b) it is clear that the actuator length is the height of 1 scissor level. Hence, by inspection, we have:

$$h = nl \tag{53}$$

$$\Rightarrow \quad \frac{dh}{dl} = n \tag{54}$$

Substituting (64) in (1) gives the required force expression, as derived in other sources as well:

$$F = n\left(L + \frac{B}{2}\right) \tag{55}$$

We will now show that we can arrive at the same result using the system derived in this paper as well. However, this is one of the rare cases when deriving from first principles is simpler.

From Fig. 10(b), we see that force is being applied on a hinge shared by a positively sloping arm and a negatively sloping arm. Hence, by the rule described in Sec. VI. A. we must consider the arm that lies entirely above the point. i.e. the positively sloping arm of the second scissor level.

It should be noted that we are not necessarily concerned with which arm the moment is directly applied to, since that is clearly the negatively sloping arm of the first scissor level. Rather, when using this system, we are concerned with the position of the point of application of force, which is, by the rule described earlier, on the positively sloping arm of the second level. This follows directly from the methodology used in the derivation.

Hence, we must use the expression for positively sloping arms:

$$F = n\left(L + \frac{B}{2}\right)\frac{\sqrt{C - \ 2(b - B)\cos\theta + D\ cos^2\theta}}{(b - B)\tan\theta - D\sin\theta} \tag{56}$$

The force is applied directly at the hinge of the arm in question. Hence, the distance between the point of application of force and the hinge immediately below that point is zero.

i.e. $a = 0$

The static end of the actuator is affixed to the origin point itself. Hence the distance of that point from the origin is zero.

i.e. $b = 0$

There is one full scissor level immediately below the point of application of force.

i.e. $i = 1$

We must now obtain the three constants, B, C and D, given that:

$a = 0$
$b = 0$
$i = 1$

The constants are given by:

$B = b(1 - a) = 0(1 - 1) = 0$
$C = b^2 + (i + a)^2 = 1$
$D = i(2a + i) = 1(0 + 1) = 1$

Substituting the values of the constants, the expression now becomes:

$$F = n\left(L + \frac{B}{2}\right)\frac{\sqrt{1 - \ 2(0)\cos\theta + \ cos^2\theta}}{(0)\tan\theta - \sin\theta} \tag{57}$$

$$\Rightarrow \quad F = n\left(L + \frac{B}{2}\right)\frac{\sqrt{(1 - \ cos^2\theta)}}{-\sin\theta} \tag{58}$$

Using the substitution $1 - cos^2\theta \equiv sin^2\theta$

$$\Rightarrow \quad F = n\left(L + \frac{B}{2}\right)\frac{\sqrt{sin^2\theta}}{-\sin\theta} \tag{59}$$

$$\Rightarrow \quad F = \pm\, n\left(L + \frac{B}{2}\right)\frac{\sin\theta}{-\sin\theta} \tag{60}$$

$$\Rightarrow \quad F = \pm\, n\left(L + \frac{B}{2}\right)\frac{\sin\theta}{\sin\theta} \tag{61}$$

$$\Rightarrow \quad F = \pm n\left(L + \frac{B}{2}\right) \tag{62}$$

Ignore negative since force in one direction cannot be negative:

$$\Rightarrow \quad F = n\left(L + \frac{B}{2}\right) \tag{63}$$

This is exactly the same expression that was obtained by inspection.

# VIII. Discussion

Current literature on an analytical description of scissor lifts either focusses on the traditional screw jack configuration, which can often be slow, or derives a separate expression for each actuator position.

This paper derives a new system to obtain the force expression for any scissor lift, and, it should be noted, the velocity ratio. This is because $\frac{dh}{dl}$, which is what was derived and then substituted into (1), is essentially the instantaneous velocity ratio of the lift as a function of the angle. Hence, the system described in this paper can be used for analysis of not only force output but also velocity ratio. This system allows one to essentially 'work backwards' to find an optimal position, by examining what values of $a$, $b$ and $i$ give the desired result. The actuator position can then be determined based on $a$, $b$ and $i$.

# IX. Conflict of Interest

The author declares that there is no conflict of interest.